\documentclass{article}

\usepackage{PRIMEarxiv}

\usepackage[utf8]{inputenc} % allow utf-8 input
\usepackage[T1]{fontenc}    % use 8-bit T1 fonts
\usepackage{hyperref}       % hyperlinks
\usepackage{url}            % simple URL typesetting
\usepackage{booktabs}       % professional-quality tables
\usepackage{amsfonts}       % blackboard math symbols
\usepackage{amsmath}        % math environments
\usepackage{nicefrac}       % compact symbols for 1/2, etc.
\usepackage{microtype}      % microtypography
\usepackage{fancyhdr}       % header
\usepackage{graphicx}       % graphics
\usepackage{multirow}       % multi-row table cells
\usepackage{subfig}         % subfigures
\usepackage{placeins}       % for \FloatBarrier
\graphicspath{{Figures/}}   % organize your images and other figures under Figures/ folder

\title{MedSR-Vision: Deep Learning Framework for Multi-Domain Medical Image Super-Resolution
\thanks{\textit{\underline{Citation}}: 
\textbf{S. Gurappa, T. Satharasi, Y. Hariprasad, and S. S. Iyengar. MedSR-Vision: Deep Learning Framework for Multi-Domain Medical Image Super-Resolution. 2025.}} 
}

\author{
  Subhash Gurappa \\
  Knight Foundation School of Computing and Information Sciences \\
  Florida International University \\
  Miami, FL 33199, USA \\
  \texttt{sg001@fiu.edu} \\
   \And
  Trivikram Satharasi \\
  University of Florida \\
  Gainesville, FL 32611, USA \\
  \texttt{t.satharasi@ufl.edu} \\
   \And
  Yashas Hariprasad \\
  Department of Computer Science \\
  California State University, East Bay \\
  Hayward, CA 94542, USA \\
  \texttt{yashas.hariprasad@csueastbay.edu} \\
   \And
  Sundararaj Sitharama Iyengar \\
  Knight Foundation School of Computing and Information Sciences \\
  Florida International University \\
  Miami, FL 33199, USA \\
  \texttt{iyengar@cs.fiu.edu} \\
}

\begin{document}
\maketitle

\begin{abstract}
Medical image super-resolution (MedSR) is essential for improving diagnostic precision across diverse imaging modalities such as MRI, CT, X-ray, Ultrasound, and Fundus imaging. Despite rapid advances in deep learning, challenges remain in preserving anatomical accuracy, maintaining perceptual quality, and generalizing across medical domains. This paper presents MedSR-Vision, a novel unified deep learning framework for evaluating and comparing super-resolution models across five modalities: Brain MRI, Chest X-ray, Renal Ultrasound, Nephrolithiasis CT, and Spine MRI, at magnification scales of $\times2$, $\times3$, and $\times4$. Three representative models namely SRCNN, SwinIR, and Real-ESRGAN are benchmarked using multiple quantitative metrics encompassing fidelity, perceptual realism, and sharpness. 

Experimental analysis demonstrates that Real-ESRGAN achieves superior perceptual quality and edge recovery at higher scales, SwinIR excels in preserving structural and diagnostic features, and SRCNN provides efficient and stable performance at lower magnifications. The results establish domain-specific insights and practical guidelines for model selection in clinical imaging workflows, offering a standardized evaluation framework for future medical image super-resolution research and deployment.
\end{abstract}

% keywords can be removed
\keywords{Medical Image Super-Resolution \and Deep Learning \and Transformer and GAN Models \and Diagnostic Image Enhancement \and Clinical Imaging Evaluation}

\section{Introduction}
\label{sec:introduction}

Medical imaging underpins diagnosis, prognosis, and treatment planning in modern healthcare. High-resolution imaging modalities such as MRI, CT, and X-ray are indispensable for detecting fine anatomical details. However, in practice, constraints such as acquisition time, radiation dose, and sensor limitations often yield low-resolution (LR) data. Enhancing such images using Super-Resolution (SR) algorithms can restore lost structural details and improve diagnostic clarity \cite{greenspan2009super,isaac2015super}. 

While conventional interpolation and reconstruction methods (e.g., bicubic, spline) produce overly smooth results \cite{isaac2015super}, deep learning based SR models have demonstrated remarkable improvements by learning complex mappings between LR and HR domains \cite{qiu2023medical,yang2023deep,elshafai2024single,umirzakova2024medical}. Yet, in medical imaging, diagnostic reliability is paramount, SR algorithms must enhance detail without introducing hallucinated artifacts that can mislead clinicians \cite{qiu2023medical,yang2023deep,elshafai2024single}. 

Our study investigates how recent deep-learning SR architectures, CNN-based (SRCNN), Transformer-based (SwinIR), and GAN-based (Real-ESRGAN), perform across multiple medical modalities and magnification scales. The aim is to develop a domain-specific evaluation benchmark, analyzing perceptual fidelity, diagnostic similarity, and sharpness metrics to guide the selection of optimal SR models for clinical applications \cite{soni2024comparative,gu2020medsrgan,yamashita2020medical}.

\section{Related Work}
\label{sec:related}

\subsection{Overview of Medical Image Super-Resolution}
Medical image super-resolution (SR) has evolved as a pivotal area of research to enhance the diagnostic quality of clinical imagery across multiple modalities such as MRI, CT, ultrasound, and retinal imaging \cite{qiu2023medical,yang2023deep,elshafai2024single,umirzakova2024medical}. Conventional interpolation-based methods (e.g., bicubic or Lanczos) often fail to preserve fine structural details and introduce blurring artifacts \cite{isaac2015super}, which critically impact diagnostic accuracy. With the rise of deep learning, data-driven SR models have achieved substantial improvements by learning complex LR--HR mappings \cite{qiu2023medical,yang2023deep,yamashita2020medical}. Qiu et al. (2023) present a comprehensive survey of CNNs, residual/dense connections, and GANs in medical SR \cite{qiu2023medical}; related reviews further detail advances in reconstruction fidelity and perceptual realism \cite{yang2023deep,elshafai2024single,umirzakova2024medical}.

Advances in medical image super-resolution intersect with AI-driven cybersecurity and digital forensics, as high-resolution medical data require secure transmission, storage, and analysis. Integrating MedSR with AI/ML-based cyber-defense frameworks enables secure imaging workflows and trustworthy clinical deployment \cite{HariprasadTCE2024,GurappaICISPD2025,IyengarFutureAIDF2025,IyengarConvergence2025,gupta2026deepfake}.

\subsection{Early Deep Learning Approaches: SRCNN and CNN-Based SR}
The early CNN paradigm for SISR established strong baselines for medical SR and spurred deeper residual/dense designs \cite{qiu2023medical,yang2023deep}. Subsequent improvements introduced residual learning, skip connections, and multi-scale designs (e.g., VDSR/EDSR/DRRN-like families) to overcome gradient degradation and improve convergence, including medical-specific residual variants \cite{qiu2021multiple}. Despite advances, SRCNN-like baselines remain useful due to simplicity, interpretability, and low computational footprint, especially in low-resource settings \cite{qiu2023medical}.

\subsection{Transformer and Attention-Based SR: SwinIR}
Transformer-based SR leverages self-attention for long-range dependency modeling, improving contextual fidelity and texture refinement relative to purely convolutional methods \cite{yang2023deep}. In medical imaging, attention mechanisms have been reported to preserve tissue morphology and subtle pathological details more effectively than classic CNNs in several settings \cite{qiu2023medical,yang2023deep}. Evidence of SwinIR's effectiveness on medical imagery has also been reported \cite{puttagunta2022swinir}.

\subsection{Real-World Degradation and GAN-Based Models: Real-ESRGAN}
GAN-based approaches model complex degradations and encourage high-frequency detail recovery via perceptual/adversarial losses \cite{qiu2023medical,yang2023deep}. Real-ESRGAN's robustness to blur, noise, and compression aligns with needs in ultrasound and CT, where acquisition artifacts are common; comparative studies across magnification scales corroborate perceptual advantages at higher $\times$ factors \cite{soni2024comparative,elshafai2024single}. Related medical SR GANs (e.g., MedSRGAN and improved GAN variants) further demonstrate clinically meaningful detail restoration \cite{gu2020medsrgan,yamashita2020medical}.

\subsection{Progressive GANs and Multi-Stage Refinement}
Mahapatra et al.\ introduce a progressive GAN for medical SR with stage-wise refinement and triplet loss, improving fine anatomical textures and aiding downstream tasks \cite{mahapatra2019image}. Such progressive/multi-stage ideas inspire hybrid pipelines in multi-domain SR frameworks.

\subsection{Comparative Medical SR Studies}
Comparative evaluations generally show CNN-based SR excels in pixel fidelity (PSNR/SSIM), while attention- and GAN-based models improve perceptual quality and texture preservation, particularly at larger scales \cite{qiu2023medical,yang2023deep,elshafai2024single,soni2024comparative,umirzakova2024medical}. Our study extends these analyses across five medical domains and three scales with a unified evaluation.

\section{Methodology}
\label{sec:methodology}

\subsection{Model Architectures}

\subsubsection{SRCNN (Baseline CNN)}
The \textbf{Super-Resolution Convolutional Neural Network (SRCNN)} family represents a classical end-to-end CNN approach for SISR and remains a key baseline in medical SR surveys \cite{qiu2023medical,yang2023deep}. It learns a direct nonlinear mapping from a low-resolution input $Y$ to a high-resolution output $X$ via three convolutional layers.

The reconstruction is expressed as:
\begin{equation}
\hat{X} = F(Y;\Theta) = f_3\left(W_3 * f_2\left(W_2 * f_1(W_1 * Y + b_1) + b_2\right) + b_3\right)
\end{equation}
where $W_i$, $b_i$ are trainable parameters, $f_i$ denotes ReLU activations, and $*$ is convolution.

Training minimizes pixel-wise \textbf{Mean Squared Error (MSE)}:
\begin{equation}
\mathcal{L}_\text{MSE} = \frac{1}{N}\sum_{i=1}^{N}\left\lVert F(Y_i;\Theta) - X_i \right\rVert^2
\end{equation}

SRCNN produces high PSNR at low scaling factors ($\times2$) but is limited by its shallow receptive field, leading to losses in fine anatomical detail; nevertheless, its low computational cost is attractive for real-time settings \cite{qiu2023medical}.

\subsubsection{SwinIR (Transformer-Based SR)}
\textbf{SwinIR (Image Restoration using Swin Transformer)} builds upon the Vision Transformer architecture by introducing shifted window attention and hierarchical feature extraction for super-resolution; transformer-based SR has been reported to improve contextual modeling and structural fidelity in medical imagery \cite{yang2023deep,qiu2023medical,puttagunta2022swinir}.

It consists of three main modules: (1) Shallow Feature Extraction, (2) Deep Feature Encoding using Swin Transformer Blocks (STBs), and (3) Reconstruction for HR synthesis.

The shallow features are extracted as:
\begin{equation}
F_0 = H_{\text{SF}}(Y)
\end{equation}

Deep features are computed through $K$ Residual Swin Transformer Blocks (RSTBs):
\begin{equation}
F_{\text{deep}} = H_{\text{RSTB}}^{K}(F_0)
\end{equation}

Each Swin Transformer Layer (STL) applies multi-head self-attention (MSA) over shifted windows:
\begin{equation}
\text{MSA}(Q,K,V) = \text{Softmax}\!\left( \frac{QK^{T}}{\sqrt{d}} + B \right)V
\end{equation}

The shifted-window mechanism improves long-range modeling while keeping computation local:
\begin{equation}
\text{SW-MSA}(x) = \text{MSA}(\text{shift}(x))
\end{equation}

The reconstruction stage generates the final SR image:
\begin{equation}
\hat{X} = H_{\text{REC}}(F_{\text{deep}} + F_0)
\end{equation}

SwinIR often employs a \textbf{Charbonnier loss}, a robust variant of $\ell_1$, which aids perceptual fidelity:
\begin{equation}
\mathcal{L}_{\text{Charb}} = \frac{1}{N}\sum_{i=1}^{N}\sqrt{(X_i - \hat{X}_i)^2 + \epsilon^2}
\end{equation}

\subsubsection{Real-ESRGAN (GAN-Based SR)}
\textbf{Real-ESRGAN} enhances ESRGAN for real-world use by reducing the domain gap between synthetic and clinical degradations via multi-stage degradation modeling and adversarial/perceptual objectives \cite{qiu2023medical,elshafai2024single,soni2024comparative,gu2020medsrgan}. 

The generator adopts Residual-in-Residual Dense Blocks (RRDBs) without batch normalization to stabilize deep residual learning. A U-Net discriminator enforces global and local fidelity to anatomical structures \cite{gu2020medsrgan}. 

The overall objective combines pixel, perceptual, and adversarial losses:
\begin{equation}
\mathcal{L}_{\text{total}}
= \lambda_1 \mathcal{L}_{\text{pixel}}
+ \lambda_2 \mathcal{L}_{\text{perc}}
+ \lambda_3 \mathcal{L}_{\text{GAN}}
\end{equation}

To better model clinical image conditions, Real-ESRGAN introduces a high-order degradation process including blur, downsampling, Gaussian noise, and compression artifacts, improving robustness to ultrasound speckle or CT photon noise \cite{elshafai2024single,yamashita2020medical}.

\subsection{Image Domains}

To comprehensively evaluate the proposed framework, three state-of-the-art super-resolution models, \textbf{SRCNN}, \textbf{SwinIR}, and \textbf{Real-ESRGAN}, were benchmarked across five medical imaging domains: \textit{Brain MRI}, \textit{Chest X-ray}, \textit{Renal Ultrasound}, \textit{Nephrolithiasis CT}, and \textit{Spine MRI}. Each domain represents a unique diagnostic scenario with distinct noise, texture, and structural complexity. Experiments were conducted on paired low- and high-resolution (LR--HR) datasets generated through controlled downsampling at scales of $\times2$, $\times3$, and $\times4$ to simulate clinically relevant degradations \cite{qiu2023medical,yang2023deep,umirzakova2024medical}. 

To evaluate resolution enhancement while preserving diagnostic quality and avoiding hallucinated details, \textit{actual medical images} were used for testing, as recommended by prior studies \cite{qiu2023medical,elshafai2024single,yamashita2020medical}. Model sensitivity to varying input sizes was also analyzed across all domains.

\begin{figure}[t]
    \centering
    \subfloat[Brain 1]{\includegraphics[width=0.30\linewidth]{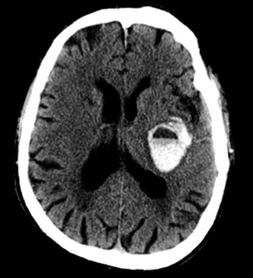}}
    \hfill
    \subfloat[Brain 2]{\includegraphics[width=0.30\linewidth]{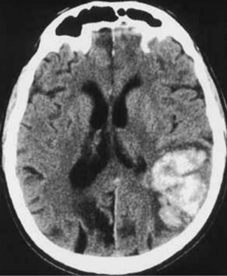}}
    \hfill
    \subfloat[Brain 3]{\includegraphics[width=0.30\linewidth]{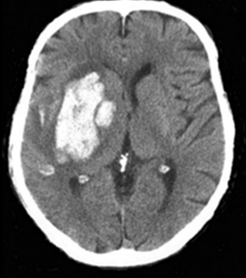}}
    \caption{Example brain CT images with various hemorrhagic presentations.}
    \label{fig:brain_examples}
\end{figure}

\subsubsection{Brain (MRI/CT)}
Brain MRI/CT imaging requires high spatial precision to capture cortical boundaries, small vessels, and lesion margins with diagnostic accuracy. As shown in Fig.~\ref{fig:brain_examples}, SR enhances subtle details (e.g., micro-hemorrhages, edema) while preserving gray--white matter contrast and avoiding hallucinated textures \cite{qiu2023medical,yang2023deep,yamashita2020medical}. Transformers maintain structural coherence, whereas GANs recover fine texture, both demanding care to prevent unrealistic artifacts \cite{qiu2023medical,elshafai2024single}.

\subsubsection{Chest (X-ray)}
\begin{figure}[t]
    \centering
    \subfloat[Chest 1]{\includegraphics[width=0.30\linewidth]{chest1.png}}
    \hfill
    \subfloat[Chest 2]{\includegraphics[width=0.30\linewidth]{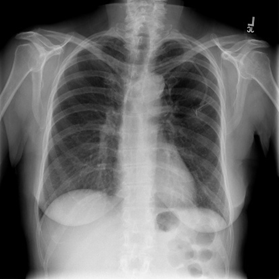}}
    \hfill
    \subfloat[Chest 3]{\includegraphics[width=0.30\linewidth]{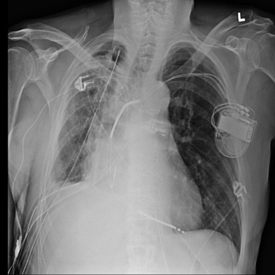}}
    \caption{Sample chest X-ray images showing lung field structural differences.}
    \label{fig:chest_examples}
\end{figure}

Chest radiography remains the most widely used imaging modality in clinical practice. Enhanced spatial resolution is essential for visualizing lung parenchyma, vascular markings, and cardiac silhouettes with diagnostic clarity, as illustrated in Fig.~\ref{fig:chest_examples}; SR methods must avoid hallucinated structures that could mislead interpretation \cite{qiu2023medical,yang2023deep,umirzakova2024medical}.

\subsubsection{Renal (Ultrasound)}
\begin{figure}[t]
    \centering
    \subfloat[Renal 1]{\includegraphics[width=0.30\linewidth]{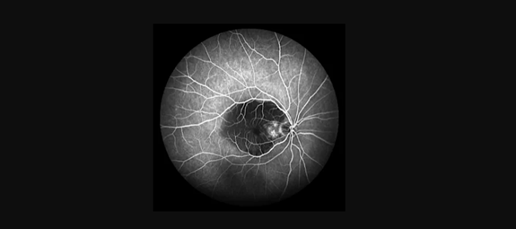}}
    \hfill
    \subfloat[Renal 2]{\includegraphics[width=0.30\linewidth]{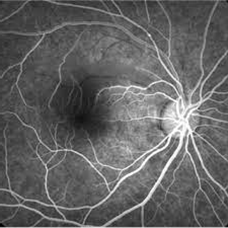}}
    \hfill
    \subfloat[Renal 3]{\includegraphics[width=0.30\linewidth]{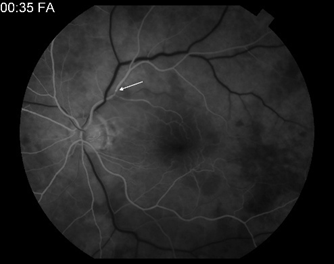}}
    \caption{Kidney ultrasound images showing renal boundaries and lesions.}
    \label{fig:renal_examples}
\end{figure}

Ultrasound imaging is constrained by speckle noise and low spatial resolution as shown in Fig.~\ref{fig:renal_examples}. SR aims to enhance anatomical visibility while preserving textural cues; GAN-based methods often show perceptual advantages in noisy modalities \cite{elshafai2024single,gu2020medsrgan}.

\subsubsection{Nephrolithiasis (CT)}
\begin{figure}[t]
    \centering
    \subfloat[CT 1]{\includegraphics[width=0.30\linewidth]{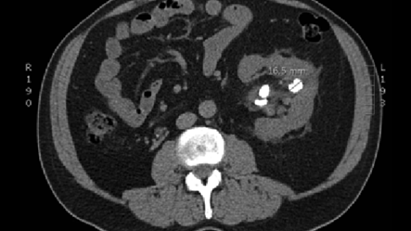}}
    \hfill
    \subfloat[CT 2]{\includegraphics[width=0.30\linewidth]{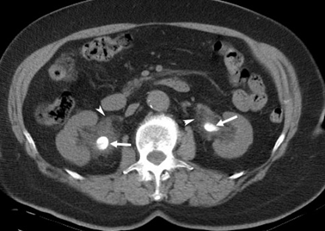}}
    \hfill
    \subfloat[CT 3]{\includegraphics[width=0.30\linewidth]{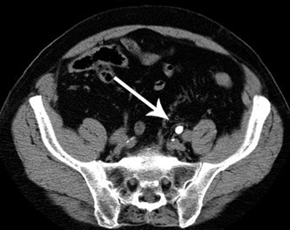}}
    \caption{Kidney CT images showing renal anatomical structures and stone visibility.}
    \label{fig:kidneyct_examples}
\end{figure}

Low-dose CT can degrade resolution and contrast-to-noise ratio, as seen in Fig.~\ref{fig:kidneyct_examples}; SR can improve edge definitions and density gradients around stone boundaries. Sharpness-oriented metrics (e.g., Laplacian variance, Tenengrad) provide complementary evidence to fidelity/perceptual metrics \cite{qiu2023medical,yamashita2020medical}.

\subsubsection{Spine (MRI)}
\begin{figure}[t]
    \centering
    \subfloat[Spine 1]{\includegraphics[width=0.30\linewidth]{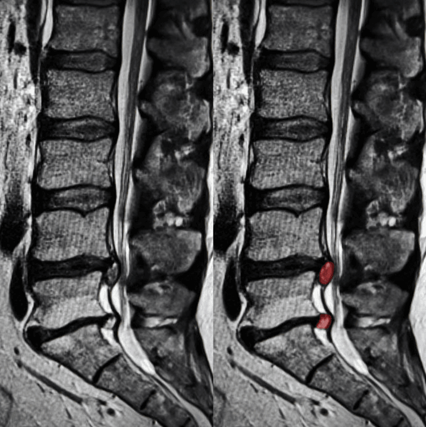}}
    \hfill
    \subfloat[Spine 2]{\includegraphics[width=0.30\linewidth]{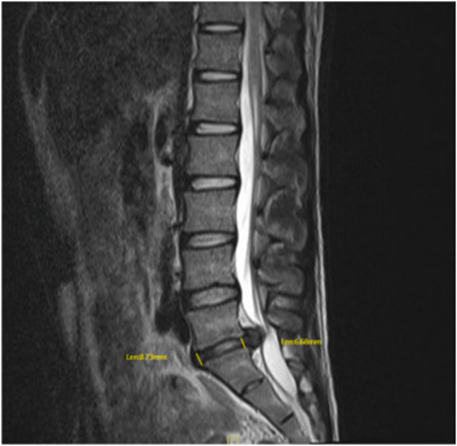}}
    \hfill
    \subfloat[Spine 3]{\includegraphics[width=0.30\linewidth]{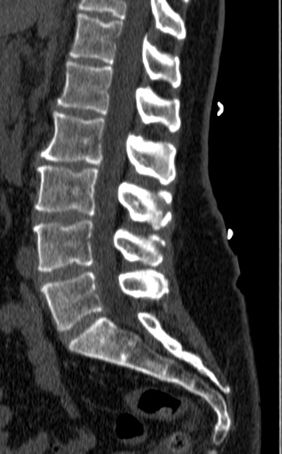}}
    \caption{Spinal images demonstrating disc bulge and vertebral variation.}
    \label{fig:spine_examples}
\end{figure}

SR can enhance edge continuity and texture sharpness in vertebral and disc regions as shown in Fig.~\ref{fig:spine_examples}; attention mechanisms tend to preserve global context alongside local detail \cite{yang2023deep,puttagunta2022swinir}.

\subsection{Evaluation Metrics}
To accurately evaluate the performance across diverse medical imaging domains, we employ sharpness, similarity, and perceptual quality metrics that together provide a multidimensional view of SR behavior \cite{qiu2023medical,yang2023deep,elshafai2024single,soni2024comparative}. These include Tenengrad Sharpness, Laplacian Variance, PSNR, SSIM, FSIM, LPIPS, and ODI Score.

\section{Experimental Setup and Results}
\label{sec:results}

\subsection{Experimental Setup}
All experiments were conducted on a high-performance computing system equipped with an NVIDIA A100 GPU (32 GB VRAM). Each model was tested using its official pretrained weights, and evaluation scripts were standardized to ensure reproducibility across all experiments, following guidance common in comparative SR studies \cite{soni2024comparative}. We also align our data handling and analytics practices with broader AI--ML experimental reporting and reproducibility considerations \cite{hariprasad2023aiml}. Workforce training in digital forensic methodologies \cite{hariprasad2026finds} further supports the responsible deployment of such AI-driven imaging systems.

\begin{table*}[!t]
\centering
\scriptsize
\setlength{\tabcolsep}{5pt}
\renewcommand{\arraystretch}{1.1}
\caption{Quantitative results for Renal Ultrasound and Chest X-ray grouped by scale.}
\label{tab:results_table1}
\begin{tabular}{@{}lllrrr@{}}
\toprule
\textbf{Image Type} & \textbf{Scale} & \textbf{Metrics} & \textbf{SRCNN} & \textbf{SwinIR} & \textbf{Real-ESRGAN}\\
\midrule
\multirow{27}{*}{Renal} 
& \multirow{9}{*}{$\times2$} & PSNR & 35.89 & 33.94 & 29.13 \\
& & SSIM & 0.97 & 0.97 & 0.73 \\
& & FSIM & 0.97 & 0.97 & 0.71 \\
& & LPIPS $\downarrow$ & 0.03 & 0.04 & 0.05 \\
& & ODI Score $\downarrow$ & 0.25 & 0.30 & 0.24 \\
& & VIF & 0.55 & 0.47 & 0.41 \\
& & Tenengrad sharpness & 1406.36 & 4420.08 & 3794.60 \\
& & Laplacian variance & 225.12 & 2810.31 & 2724.06 \\
& & Brenner & 49.17 & 199.71 & 170.32 \\
\cmidrule{2-6}
& \multirow{9}{*}{$\times3$} & PSNR & 37.03 & -- & 29.09 \\
& & SSIM & 0.98 & -- & 0.73 \\
& & FSIM & 0.98 & -- & 0.71 \\
& & LPIPS $\downarrow$ & 0.07 & -- & 0.06 \\
& & ODI Score $\downarrow$ & 0.03 & -- & 0.15 \\
& & VIF & 0.50 & -- & 0.43 \\
& & Tenengrad sharpness & 305.46 & -- & 2705.92 \\
& & Laplacian variance & 15.74 & -- & 1166.27 \\
& & Brenner & 9.27 & -- & 115.35 \\
\cmidrule{2-6}
& \multirow{9}{*}{$\times4$} & PSNR & 39.07 & 29.39 & 29.07 \\
& & SSIM & 0.98 & 0.91 & 0.72 \\
& & FSIM & 0.98 & 0.92 & 0.71 \\
& & LPIPS $\downarrow$ & 0.03 & 0.11 & 0.07 \\
& & ODI Score $\downarrow$ & 0.00 & 0.16 & 0.12 \\
& & VIF & 0.54 & 0.24 & 0.47 \\
& & Tenengrad sharpness & 80.79 & 3770.90 & 1934.68 \\
& & Laplacian variance & 7.29 & 3259.60 & 554.53 \\
& & Brenner & 2.36 & 182.31 & 80.13 \\
\midrule
\multirow{27}{*}{Chest} 
& \multirow{9}{*}{$\times2$} & PSNR & 40.31 & 38.92 & 40.31 \\
& & SSIM & 0.99 & 0.95 & 0.95 \\
& & FSIM & 0.99 & 0.95 & 0.95 \\
& & LPIPS $\downarrow$ & 0.05 & 0.12 & 0.13 \\
& & ODI Score $\downarrow$ & 0.11 & 0.33 & 0.32 \\
& & VIF & 0.77 & 0.69 & 0.74 \\
& & Tenengrad sharpness & 206.74 & 366.35 & 244.81 \\
& & Laplacian variance & 41.83 & 241.04 & 162.26 \\
& & Brenner & 8.06 & 16.55 & 10.71 \\
\cmidrule{2-6}
& \multirow{9}{*}{$\times3$} & PSNR & 40.47 & -- & 40.09 \\
& & SSIM & 0.99 & -- & 0.95 \\
& & FSIM & 0.99 & -- & 0.95 \\
& & LPIPS $\downarrow$ & 0.07 & -- & 0.15 \\
& & ODI Score $\downarrow$ & 0.00 & -- & 0.15 \\
& & VIF & 0.72 & -- & 0.71 \\
& & Tenengrad sharpness & 74.52 & -- & 146.12 \\
& & Laplacian variance & 17.03 & -- & 70.56 \\
& & Brenner & 2.91 & -- & 6.54 \\
\cmidrule{2-6}
& \multirow{9}{*}{$\times4$} & PSNR & 42.60 & 35.36 & 40.09 \\
& & SSIM & 0.99 & 0.91 & 0.96 \\
& & FSIM & 0.99 & 0.91 & 0.96 \\
& & LPIPS $\downarrow$ & 0.06 & 0.25 & 0.19 \\
& & ODI Score $\downarrow$ & 0.00 & 0.27 & 0.32 \\
& & VIF & 0.66 & 0.38 & 0.70 \\
& & Tenengrad sharpness & 39.42 & 333.52 & 99.85 \\
& & Laplacian variance & 15.18 & 336.56 & 44.18 \\
& & Brenner & 1.55 & 15.62 & 4.47 \\
\bottomrule
\end{tabular}
\end{table*}

\begin{table*}[!t]
\centering
\scriptsize
\setlength{\tabcolsep}{5pt}
\renewcommand{\arraystretch}{1.1}
\caption{Quantitative results for Brain MRI and Nephrolithiasis CT grouped by scale.}
\label{tab:results_table2}
\begin{tabular}{@{}lllrrr@{}}
\toprule
\textbf{Image Type} & \textbf{Scale} & \textbf{Metrics} & \textbf{SRCNN} & \textbf{SwinIR} & \textbf{Real-ESRGAN}\\
\midrule
\multirow{27}{*}{Brain} 
& \multirow{9}{*}{$\times2$} & PSNR & 28.49 & 23.39 & 25.14 \\
& & SSIM & 0.97 & 0.77 & 0.76 \\
& & FSIM & 0.97 & 0.77 & 0.76 \\
& & LPIPS $\downarrow$ & 0.11 & 0.33 & 0.42 \\
& & ODI Score $\downarrow$ & 0.13 & 0.27 & 0.27 \\
& & VIF & 0.76 & 0.30 & 0.25 \\
& & Tenengrad sharpness & 1997.12 & 9739.12 & 6437.11 \\
& & Laplacian variance & 76.11 & 10506.52 & 6434.58 \\
& & Brenner & 69.22 & 509.76 & 315.79 \\
\cmidrule{2-6}
& \multirow{9}{*}{$\times3$} & PSNR & 27.27 & -- & 26.05 \\
& & SSIM & 0.95 & -- & 0.73 \\
& & FSIM & 0.95 & -- & 0.74 \\
& & LPIPS $\downarrow$ & 0.13 & -- & 0.45 \\
& & ODI Score $\downarrow$ & 0.03 & -- & 0.13 \\
& & VIF & 0.57 & -- & 0.19 \\
& & Tenengrad sharpness & 1034.86 & -- & 4951.13 \\
& & Laplacian variance & 33.11 & -- & 4907.33 \\
& & Brenner & 34.85 & -- & 242.91 \\
\cmidrule{2-6}
& \multirow{9}{*}{$\times4$} & PSNR & 27.91 & 21.90 & 24.96 \\
& & SSIM & 0.93 & 0.69 & 0.74 \\
& & FSIM & 0.93 & 0.69 & 0.74 \\
& & LPIPS $\downarrow$ & 0.12 & 0.47 & 0.45 \\
& & ODI Score $\downarrow$ & 0.00 & 0.08 & 0.08 \\
& & VIF & 0.49 & 0.15 & 0.15 \\
& & Tenengrad sharpness & 617.29 & 7504.69 & 4350.47 \\
& & Laplacian variance & 23.13 & 8653.97 & 3644.17 \\
& & Brenner & 21.13 & 408.69 & 211.92 \\
\midrule
\multirow{27}{*}{Nephrolithiasis} 
& \multirow{9}{*}{$\times2$} & PSNR & 27.39 & 27.53 & 26.16 \\
& & SSIM & 0.94 & 0.83 & 0.77 \\
& & FSIM & 0.94 & 0.83 & 0.77 \\
& & LPIPS $\downarrow$ & 0.13 & 0.24 & 0.34 \\
& & ODI Score $\downarrow$ & 0.19 & 0.20 & 0.17 \\
& & VIF & 0.63 & 0.43 & 0.30 \\
& & Tenengrad sharpness & 2409.22 & 6968.46 & 7940.53 \\
& & Laplacian variance & 225.02 & 3810.47 & 5744.80 \\
& & Brenner & 87.00 & 319.56 & 370.57 \\
\cmidrule{2-6}
& \multirow{9}{*}{$\times3$} & PSNR & 27.87 & -- & 26.13 \\
& & SSIM & 0.93 & -- & 0.78 \\
& & FSIM & 0.93 & -- & 0.78 \\
& & LPIPS $\downarrow$ & 0.15 & -- & 0.43 \\
& & ODI Score $\downarrow$ & 0.02 & -- & 0.13 \\
& & VIF & 0.48 & -- & 0.30 \\
& & Tenengrad sharpness & 1006.11 & -- & 5383.86 \\
& & Laplacian variance & 46.11 & -- & 3321.78 \\
& & Brenner & 36.11 & -- & 239.76 \\
\cmidrule{2-6}
& \multirow{9}{*}{$\times4$} & PSNR & 30.65 & 26.90 & 26.08 \\
& & SSIM & 0.93 & 0.83 & 0.79 \\
& & FSIM & 0.93 & 0.83 & 0.79 \\
& & LPIPS $\downarrow$ & 0.12 & 0.45 & 0.10 \\
& & ODI Score $\downarrow$ & 0.01 & 0.11 & 0.10 \\
& & VIF & 0.46 & 0.34 & 0.23 \\
& & Tenengrad sharpness & 451.91 & 3496.23 & 4107.04 \\
& & Laplacian variance & 23.36 & 1393.72 & 2132.35 \\
& & Brenner & 15.75 & 150.89 & 177.43 \\
\bottomrule
\end{tabular}
\end{table*}

\begin{table*}[!t]
\centering
\scriptsize
\setlength{\tabcolsep}{5pt}
\renewcommand{\arraystretch}{1.1}
\caption{Quantitative results for Spine MRI grouped by scale.}
\label{tab:results_table3}
\begin{tabular}{@{}lllrrr@{}}
\toprule
\textbf{Image Type} & \textbf{Scale} & \textbf{Metrics} & \textbf{SRCNN} & \textbf{SwinIR} & \textbf{Real-ESRGAN}\\
\midrule
\multirow{27}{*}{Spine} 
& \multirow{9}{*}{$\times2$} & PSNR & 33.45 & 21.98 & 20.55 \\
& & SSIM & 0.97 & 0.65 & 0.59 \\
& & FSIM & 0.97 & 0.65 & 0.59 \\
& & LPIPS $\downarrow$ & 0.07 & 0.34 & 0.66 \\
& & ODI Score $\downarrow$ & 0.23 & 0.19 & 0.16 \\
& & VIF & 0.73 & 0.32 & 0.28 \\
& & Tenengrad sharpness & 4355.66 & 21613.18 & 22669.63 \\
& & Laplacian variance & 251.95 & 18489.04 & 25647.10 \\
& & Brenner & 195.27 & 1083.01 & 127.86 \\
\cmidrule{2-6}
& \multirow{9}{*}{$\times3$} & PSNR & 29.93 & -- & 20.54 \\
& & SSIM & 0.92 & -- & 0.54 \\
& & FSIM & 0.92 & -- & 0.54 \\
& & LPIPS $\downarrow$ & 0.12 & -- & 0.54 \\
& & ODI Score $\downarrow$ & 0.02 & -- & 0.09 \\
& & VIF & 0.55 & -- & 0.23 \\
& & Tenengrad sharpness & 1959.77 & -- & 17812.27 \\
& & Laplacian variance & 64.62 & -- & 16007.28 \\
& & Brenner & 86.83 & -- & 927.95 \\
\cmidrule{2-6}
& \multirow{9}{*}{$\times4$} & PSNR & 27.81 & 20.21 & 20.48 \\
& & SSIM & 0.90 & 0.45 & 0.52 \\
& & FSIM & 0.90 & 0.45 & 0.52 \\
& & LPIPS $\downarrow$ & 0.13 & 0.66 & 0.59 \\
& & ODI Score $\downarrow$ & -0.02 & 0.05 & 0.09 \\
& & VIF & 0.50 & 0.17 & 0.19 \\
& & Tenengrad sharpness & 1011.56 & 17241.08 & 15226.67 \\
& & Laplacian variance & 36.46 & 17067.13 & 11423.32 \\
& & Brenner & 45.03 & 915.92 & 747.86 \\
\bottomrule
\end{tabular}
\end{table*}

\subsection{Results and Discussion}

The comparative evaluation, shown in Tables~\ref{tab:results_table1}, \ref{tab:results_table2}, and \ref{tab:results_table3} across five medical imaging domains reveals differentiated strengths. GAN-based methods often score higher on perceptual/sharpness indicators, while transformer approaches tend to preserve structural similarity, in line with broader observations in medical SR surveys \cite{qiu2023medical,yang2023deep,elshafai2024single}. Our scaling trends also mirror findings from magnification-focused comparisons \cite{soni2024comparative}.

\subsubsection{Domain-Specific Analysis}
Analysis of Tables~\ref{tab:results_table1}, \ref{tab:results_table2}, and \ref{tab:results_table3} reveals that SR model effectiveness is strongly anatomy- and modality-dependent. Different models exhibit strengths tailored to specific clinical imaging characteristics. This subsection discusses the key domain-specific trends and the underlying factors contributing to model performance variations.

\paragraph{Brain MRI.} Real-ESRGAN yields high sharpness while SwinIR maintains superior SSIM; this echoes the fidelity--perception trade-off noted in prior reviews \cite{qiu2023medical,yang2023deep,umirzakova2024medical}.

\paragraph{Renal Ultrasound.} Robust perceptual enhancement aligns with GAN advantages in noisy modalities \cite{elshafai2024single,gu2020medsrgan}.

\paragraph{Chest X-ray.} Attention-driven structure preservation is consistent with transformer observations \cite{yang2023deep,puttagunta2022swinir}.

\paragraph{Nephrolithiasis CT.} Sharpness improvements complement PSNR/SSIM, supporting multi-metric evaluation \cite{qiu2023medical,yamashita2020medical}.

\paragraph{Spine MRI.} Structural consistency benefits from attention-based context modeling \cite{yang2023deep,puttagunta2022swinir}.

\subsubsection{Scale and Metric Dependency}

\paragraph{Scale $\times2$.} High PSNR consistency with low hallucination risk mirrors classic CNN behavior \cite{qiu2023medical}.

\paragraph{Scale $\times3$.} Perceptual robustness becomes more critical; GAN-based approaches often maintain texture under noise \cite{elshafai2024single,gu2020medsrgan}.

\paragraph{Scale $\times4$.} Attention models preserve structure while GANs recover high-frequency detail; similar trends are reported in comparative literature \cite{soni2024comparative,puttagunta2022swinir}.

\subsubsection{Practical Implications}
Model selection should be domain- and goal-specific, a theme emphasized across medical SR reviews \cite{qiu2023medical,yang2023deep}. A hybrid SR pipeline, attention-first for structure, GAN refinement for detail, is consistent with progressive/multi-stage insights from \cite{mahapatra2019image,gu2020medsrgan}.

\section{Conclusion}
\label{sec:conclusion}

This paper introduces a novel integrated system to evaluate deep learning--based super-resolution models across multiple medical imaging modalities and magnification levels. The study highlights that GAN- and Transformer-based approaches offer complementary strengths: Real-ESRGAN demonstrates superior perceptual quality and edge recovery, SwinIR excels in preserving structural integrity and diagnostic details, and SRCNN remains efficient for low-scale or real-time applications. These insights emphasize that model selection should be guided by the diagnostic context and computational constraints of the clinical workflow.

From a practical standpoint, attention-based models are ideal for structure-sensitive imaging such as spine MRI or chest X-rays, while GAN-based methods perform best in high-detail modalities like CT and ultrasound. A hybrid approach combining both can achieve an effective balance between perceptual sharpness and diagnostic reliability.

Future directions include developing adaptive and hybrid architectures that maintain structural accuracy, minimize hallucinated details, and ensure model generalization across scanners and clinical environments. Overall, MedSR-Vision establishes a standardized and scalable foundation for benchmarking, guiding, and deploying super-resolution models in real-world medical imaging practice.

\section*{Acknowledgments}
The authors extend their sincere gratitude to all individuals who contributed through valuable discussions in the early stages of this work.

%Bibliography
\bibliographystyle{unsrt}

\end{document}